\documentclass[letterpaper]{article} % DO NOT CHANGE THIS
\usepackage{aaai2026}  % DO NOT CHANGE THIS
\usepackage{times}  % DO NOT CHANGE THIS
\usepackage{helvet}  % DO NOT CHANGE THIS
\usepackage{courier}  % DO NOT CHANGE THIS
\usepackage[hyphens]{url}  % DO NOT CHANGE THIS
\usepackage{graphicx} % DO NOT CHANGE THIS
\usepackage{natbib}  % DO NOT CHANGE THIS AND DO NOT ADD ANY OPTIONS TO IT
\usepackage{caption} % DO NOT CHANGE THIS AND DO NOT ADD ANY OPTIONS TO IT
\usepackage{algorithm}
\usepackage{algorithmic}

\usepackage{multirow} 
\usepackage{colortbl}
\usepackage{amsfonts}
\usepackage{amsmath}
\usepackage{booktabs}
\definecolor{mygray}{gray}{.9}

\def\name{DVA}

\usepackage{newfloat}
\usepackage{listings}
\DeclareCaptionStyle{ruled}{labelfont=normalfont,labelsep=colon,strut=off} % DO NOT CHANGE THIS
\floatstyle{ruled}
\newfloat{listing}{tb}{lst}{}
\floatname{listing}{Listing}
\title{Fine-grained Image Retrieval via Dual-Vision Adaptation}
\author{
	Xin Jiang\textsuperscript{\rm 1}\thanks{Equal contribution.},
	Meiqi Cao\textsuperscript{\rm 1}\footnotemark[1],
	Hao Tang\textsuperscript{\rm 2},
        Fei Shen\textsuperscript{\rm 3},
	Zechao Li\textsuperscript{\rm 1}\thanks{Corresponding author.}
}

\affiliations{
    \textsuperscript{\rm 1}Nanjing University of Science and Technology, China\\
    \textsuperscript{\rm 2}Centre for Smart Health, Hong Kong Polytechnic University, China\\
    \textsuperscript{\rm 3}National University of Singapore, Singapore\\
    \{zechao.li\}@njust.edu.cn
}
\usepackage{bibentry}
\begin{document}

\maketitle

\begin{abstract}
Fine-Grained Image Retrieval~(FGIR) faces challenges in learning discriminative visual representations to retrieve images with similar fine-grained features.
Current leading FGIR solutions typically follow two regimes: enforce pairwise similarity constraints in the semantic embedding space, or incorporate a localization sub-network to fine-tune the entire model. 
However, such two regimes tend to overfit the training data while forgetting  the knowledge gained from large-scale pre-training, thus reducing their generalization ability.
In this paper, we propose a Dual-Vision Adaptation (DVA) approach for FGIR, which guides the frozen pre-trained model to perform FGIR through collaborative sample and feature adaptation.  
Specifically, we design Object-Perceptual Adaptation, which modifies input samples to help the pre-trained model perceive critical objects and elements within objects that are helpful for category prediction.
Meanwhile, we propose In-Context Adaptation, which introduces a small set of parameters for feature adaptation without modifying the pre-trained parameters. This makes the FGIR task using these adapted features closer to the task solved during the pre-training. 
Additionally, to balance retrieval efficiency and performance, we propose Discrimination Perception Transfer to transfer the discriminative knowledge in the object-perceptual adaptation to the image encoder using the knowledge distillation mechanism.
Extensive experiments show that DVA performs well on three fine-grained datasets.
\end{abstract}

\section{Introduction} \label{s1}
\begin{figure}
    \centering
    \includegraphics[width=0.9\linewidth]{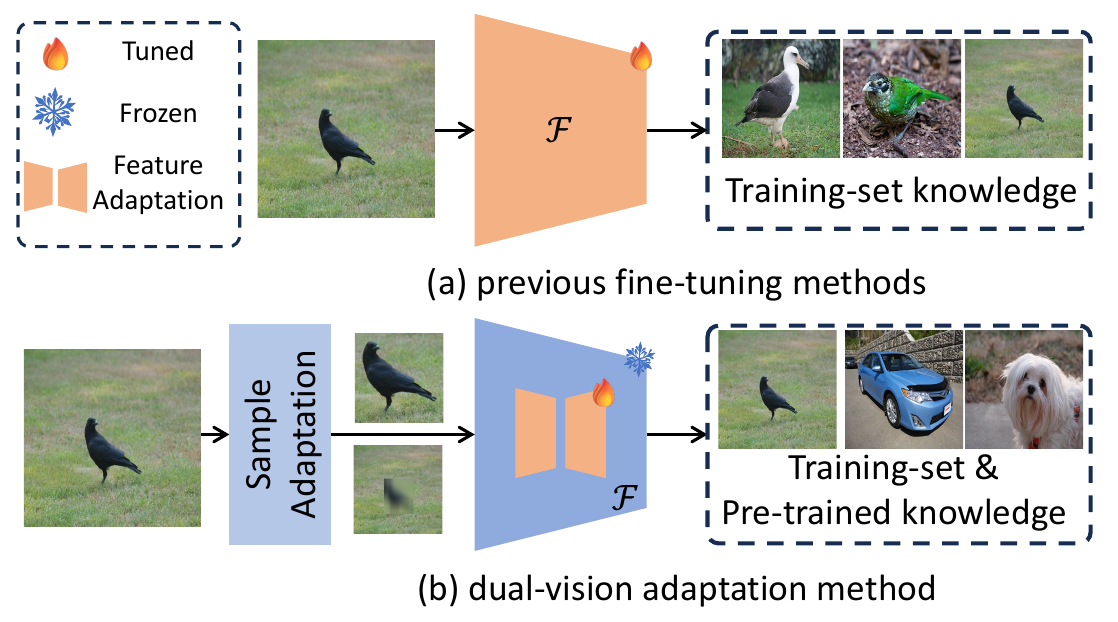}
    \caption{(a) previous fine-tuning methods. (b) our dual-vision adaptation method. Our approach designs the collaborative sample and feature adaptation to exploit category-specific differences. This dual strategy enables model to sustain broad representation capabilities from pre-training data while dynamically adjusting its adaptability to fine-grained data.}
    \label{fig:fig1}
\end{figure}
Unlike general image retrieval, fine-grained image retrieval~(FGIR) attempts to retrieve images belonging to the same subcategory as the query image from a database within a broader meta category (\emph{i.e.}, birds, cars)~\cite{WeiLWZ17}. 
It has been extensively applied across various domains, including intelligent transportation~\cite{ramdurai2025large} and biodiversity monitoring~\cite{vendrow2024inquire}. 
However, FGIR presents significant challenges due to two inherent difficulties: i) subtle visual distinctions between different categories, and ii) significant appearance variations within the same category. This task usually requires models to simultaneously localize discriminative regions and identify minute visual differences, creating a dual optimization dilemma.
Therefore, the key to FGIR is learning discriminative and generalizable embeddings to identify visually similar objects accurately.
\begin{sloppypar}
Conventional FGIR methods typically adopt two primary strategies: encoding-based~\cite{pnca,pnca++,Hyp-ViT,hist} and localization-based~\cite{ZhengJSWHY18,frpt,WeiLWZ17,wang2022coarse}.
Encoding-based methods primarily learn image-level features but struggle to suppress background and irrelevant information interference. 
In contrast, localization-based methods enhance feature extraction within the encoder to capture subtle differences among subcategories, often by focusing on distinctive object regions.
Although both strategies capture discriminative embedding learning for fine-grained retrieval, their dependence on full-parameter fine-tuning introduces prohibitive computational costs and limits cross-task generalization, as shown in Fig.~\ref{fig:fig1}(a).
Simultaneously, fine-tuning the entire model destroys the knowledge in the large-scale pre-trained parameters and may lead to suboptimal convergence, which ultimately limits retrieval performance~\cite{frpt}.
This raises the question: can discriminative representations be learned for FGIR tasks without fine-tuning the model’s pre-trained parameters? 
\end{sloppypar} 

Recent advancements in parameter-efficient fine-tuning~(PEFT) techniques have demonstrated the feasibility of learning representations while introducing a limited number of new parameters to adapt a frozen pre-trained model to downstream tasks.
Its core idea is to redesign downstream tasks to align closely with those addressed during pre-training, ensuring that large-scale pre-trained knowledge is effectively retained. 
Owing to PEFT, large-scale pre-trained visual-language models~(\emph{e.g.,} CLIP~\cite{clip}) have achieved impressive results in various visual tasks.
However, existing PEFT methods~\cite{zhou2022cocoop,zhou2022coop} are designed for multimodal models to capture high-level category semantics rather than subtle subcategory differences, so directly applying existing PEFT techniques to FGIR without task-specific adaptation tends to be suboptimal, due to their limited focus on capturing subtle intra-class variations.
Therefore, it is crucial to develop an efficient fine-tuning strategy specifically designed for fine-grained visual models, as this can prevent suboptimal convergence that may occur when fine-tuning the entire FGIR model.

To this end, we design a novel adaptation framework for FGIR, which efficiently unleashes the power of pre-trained model, as shown in Fig.~\ref{fig:fig1}(b).
%
%c 
Technically, we propose Dual-Vision Adaptation (DVA) to resolve the conflict between preserving pre-trained knowledge and acquiring fine-grained discriminative power. By maintaining the frozen backbone, DVA establishes category-sensitive perception through sample-feature co-adaptation, inherently avoiding the optimization dilemma of conventional full fine-tuning.
Specifically, the Object-Perceptual Adaption~(OPA) is a proposed sample-adapted strategy for the visual foundation model, encouraging the pre-trained model to capture critical object regions and locate subclass-specific differences. 
Meanwhile, as a feature adaptation process, the proposed In-Context Adaption~(ICA) aims to focus on fine-grained features that contribute to category prediction through lightweight learnable parameters. In this way, ICA can adapt the feature content in a direction that is conducive to category recognition, which makes the FGIR task with such adapted features close to the task solved during the original pre-training.
Nevertheless, a non-negligible problem is that the integration between the visual foundation model and the frozen pre-trained backbone introduces prohibitive computational overhead during inference. 
To address this critical efficiency bottleneck while retaining fine-grained discriminative power, we devise a Discriminative Perceptual Transfer~(DPT) module that transfers the discriminative knowledge from OPA to the backbone via a distillation mechanism.

In summary, the primary contributions of this work are as follows:

%c
\begin{itemize}
    \item We propose a dual vision adaptation framework to enable the frozen pre-trained model to capture subtle sub-category differences without forgetting the pre-trained knowledge, balancing accuracy and efficiency.
    \item We propose a dynamic sample and feature co-adaptation strategy and a distillation-based sample perception transfer module to bridge the gap between high-level semantics and fine-grained distinctions, which are realized through the collaborative modification of samples to features.
    \item With only 0.68\% of the tunable parameters compared to full fine-tuning, DVA achieves competitive performance on three fine-grained datasets.
\end{itemize}
\section{Related Works}
\subsection{Fine-Grained Image Retrieval}
Existing methods for fine-grained image retrieval~(FGIR) can be categorized as encoding-based or localization-based.
The encoding-based methods~\cite{pnca,pnca++,hist,Hyp-ViT,jiang2024dvf,jiang2024global,tang2020blockmix} aim to learn an embedding space in which samples of a similar subcategory are attracted and samples of different subcategories are repelled.
The methods can be decomposed into roughly two components: the image encoder maps images into an embedding space, and the metric method ensures that samples from the same subcategories are grouped closely, while samples from different subcategories are separated.
While these studies have achieved significant achievements, they primarily concentrate on optimizing image-level features that include numerous noisy and non-discriminatory details.
Therefore, the localization-based methods~\cite{HanGZZ18,TangLPT20,frpt,WeiLWZ17,moskvyak2021keypoint,tang2022learning} are proposed, which focus on training a subnetwork for locating discriminative regions or devising an effective strategy for extracting attractive object features to facilitate the retrieval task.
Unlike these approaches, our method considers the specific characteristics of the FGIR tasks, offering guidance for designing high-performance retrieval models.

\subsection{Parameter-Efficient Fine Tuning}
Parameter-efficient fine-tuning~\cite{hu2022lora} in NLP reformulates downstream tasks as language modeling problems, allowing pre-trained language models to adapt more efficiently to new tasks. As a result, these techniques are now widely used in various NLP applications, including language understanding and generation.
Recently, parameter-efficient fine-tuning has also been applied to multimodal computer vision~\cite{zhou2022coop,xing2025inv,mpfgvc,tang2025connecting,cao2024eventcrab,gao2025styleshot}.
However, existing methods primarily enhance language-based models, making them unsuitable for direct application to pre-trained vision models.
Additionally, some recent approaches~\cite{chen2022adaptformer,gao2025faceshot} introduce a small number of learnable parameters to guide pre-trained models in general vision tasks. 
However, their fine-tuning strategies focus on capturing category-level semantics rather than the fine-grained visual differences needed to distinguish similar objects. 
To address this limitation, we propose a parameter-efficient DVA that incorporates sample and feature adaptation, enabling the frozen pre-trained vision model to perform FGIR tasks effectively.
\begin{figure*}[t]
    \centering
    \includegraphics[width=0.85\textwidth]{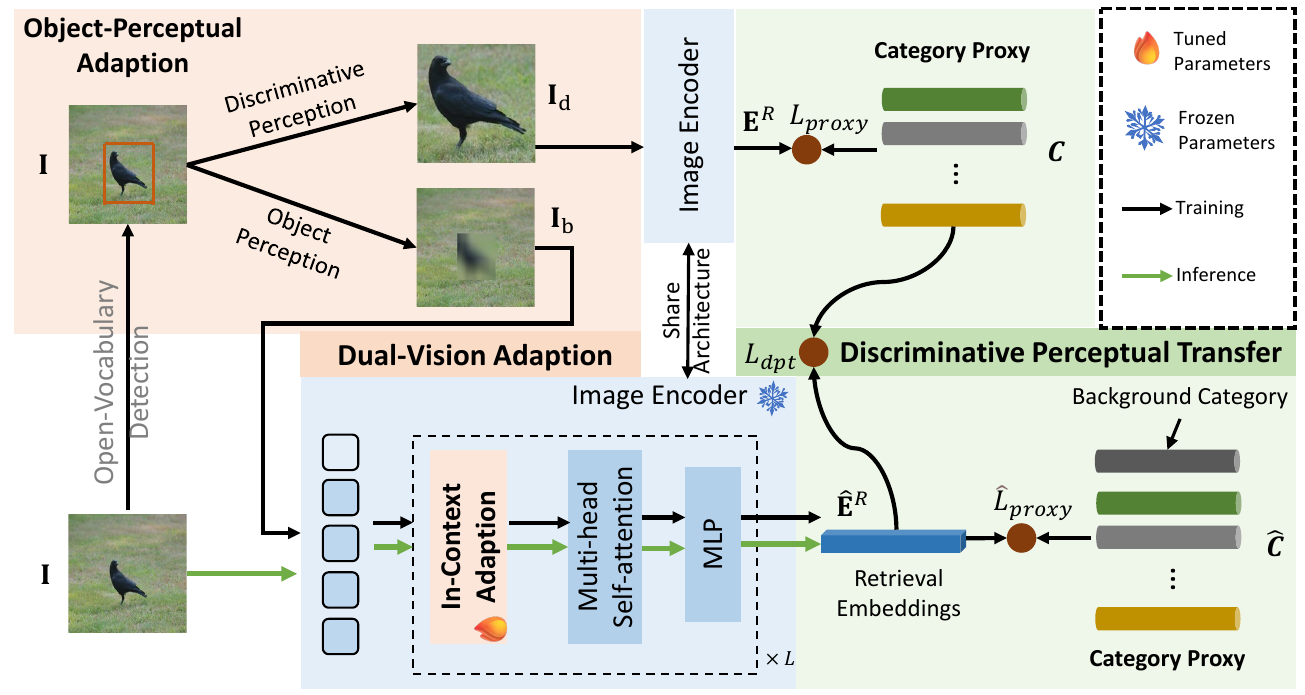}
    \caption{DVA consists of three essential modules: the Object-Perceptual Adaptation module to enhance the encoder’s ability to focus on discriminative object regions, the In-Context Adaptation module to dynamically refine fine-grained features while suppressing irrelevant background retained in frozen representations, and the Discriminative Perceptual Transfer module to distill discriminative awareness into the encoder, enabling auxiliary-free inference while preserving pre-trained knowledge.}
    \label{fig:overview}
\end{figure*}

\subsection{Knowledge Distillation}
Knowledge Distillation~(KD) is dedicated to compressing the informative knowledge from a large and computationally expensive model~(\emph{i.e,} teacher model) to a small and computationally efficient model~(\emph{i.e,} student model)~\cite{hinton2015distilling,csdnet,cao2025exploiting}. Given the strong knowledge transfer ability, it is widely applied to natural language processing, computer vision and other fields~\cite{wang2021knowledge,tang2023m3net,tang2024divide}. Most classification-based KD approaches explore enhancing the student model by mimicking the teacher model's predictions or output distribution. Additionally, researchers have delved into exploring KD for image retrieval by leveraging distances between samples, such as learning to rank~\cite{chen2018darkrank} and regression on quantities involving one or more pairs, like distances~\cite{pleor} or angles~\cite{park2019relational}. In this paper, we transfer discriminative knowledge that aids fine-grained image retrieval via proxy features.
\section{Method}
\subsection{Dual-Vision Adaptation}

\subsubsection{Object-Perceptual Adaption.} \label{opa}
Subtle yet discriminative discrepancies are widely recognized to be significant for fine-grained understanding~\cite{csdnet,frpt,cao2024eventcrab}.
However, the vanilla ViT model is originally designed to identify different species (\emph{e.g.}, cats, dogs, and birds), rather than to exploit subtle differences between subcategories within a species.
Therefore, without task-specific adaptation, the vanilla ViT has limited focus on capturing subtle intra-class variations.
To alleviate this situation, we design an Object-Perceptual Adaption~(OPA) that only modifies the visual input of the image encoder~(ViT) to assist the model in perceiving discriminative differences in fine-grained images.
Specifically, OPA consists of two components: discriminative perception and object perception. The former enhances category discrimination by focusing on foreground regions identified by the visual foundation model, while the latter improves object perception by utilizing background information separated by the same model.

\textbf{Discriminative Perception.} For fine-grained understanding, there have been many works~\cite{xing2024csgo,jiang2024dvf} demonstrating that the background can perturb the model's perception of discriminative regions.
Therefore, we design discriminative perception to help the model focus on objects in the image, thereby exploring discriminative features through critical regions.
We use visual foundation model to extract object regions from training images, keep the core object foreground, obtaining the discriminative image $\mathbf{I}_\mathrm{d}$.
Specifically, each training image along with the super-class name~(\emph{e.g.,} bird for CUB-200-2011~\cite{cub}) of the image is fed to open-vocabulary detector  GroundingDINO~\cite{grounding-dino} to obtain the bounding box that contains the foreground object of the category. 
Therefore, there is no risk of information leakage for FGIR.
Then, the region containing the foreground object is extracted from the image as the discriminative image~$\mathbf{I}_\mathrm{d}$.
To ensure objects are fully contained in the image and free from deformation, we filter out low-confidence detection results, and finally apply padding on the short sides of the image to preserve the original shape of the object.

\textbf{Object Perception.} 
In fine-grained tasks, the background often provides contextual information that is complementary to the foreground and can help distinguish similar objects. For example, the foreground features of the ``Red-bellied Tit" and the ``Golden Pheasant" are similar, but the former often appears in bushes, while the latter often appears in rocky areas.
Therefore, we aim to extract background regions from training images and establish an explicit background category to enhance model discrimination.
Specifically, we reuse the detection boxes of foreground objects obtained from discriminative perception. Unlike previous use, we apply a mean filter to blur the image region within the bounding box, resulting in a background image that excludes object information.
For those training images whose foreground objects occupy the majority of the whole image region, background images would contain much less background information after blurring foreground regions. 
Therefore, only when the area of the foreground region is smaller than a predefined proportion $\alpha \%$ (\emph{e.g.}, 50$\%$) of the whole training image, the training image is used to construct the background category image $\mathbf{I}_\mathrm{b}$ as described above.

\subsubsection{In-Context Adaptation.}
Conventional methods~\cite{Hyp-ViT,csdnet,cao2024adafpp} usually require fine-tuning the entire network, but this approach ignores the large-scale pre-trained knowledge present in the pre-trained parameters.
We hypothesize that FGIR on downstream datasets can be learned and understood by specific modules within the network.
To verify this hypothesis, we conducte an in-depth analysis of the fine-tune preferences of ViT.

We first divide the trainable parameters of transformer layers into three parts: the attention projector, output projector and MLP.
Subsequently, we separately fine-tune these three parts and ViT as a whole and evaluate on the CUB-200-2011~\cite{cub} dataset.
As shown in Table~\ref{tab:analysis}, we found that by fine-tuning only specific modules within ViT, it is possible to achieve performance comparable to that of fine-tuning the entire model.
Meanwhile, fine-tuning the attention projector yields the best performance. This is intuitive, as the self-attention mechanism is central to ViT, and the attention projector enhances its ability to encode contextual information.
This experiment further validates our hypothesis that the FGIR task can be learned by a specific network module, namely attention projector.

Based on this insight, we attempt to design a in-context adaptation module with tune only the attention projector of the ViT. 
However, this approach still destroys the knowledge in the pre-trained parameters and still requires learning a large number of parameters~(22.5M). To address this issue, we keep the attention projector frozen and propose a lighter-weight learnable in-context adaptation module in parallel with the attention projector to learn the specific knowledge required for FGIR.
Specifically, the in-context adaptation module is designed to be a bottleneck structure for limiting the number of parameters purpose, which includes a down-projection layer with parameters $\mathbf{W}_{down} \in \mathbb{R}^{D \times d}$, an up-projection layer with parameters $\mathbf{W}_{up} \in \mathbb{R}^{d \times D}$, in this work, we set $d=16$.
For a specific input feature $\mathbf{x}$, the in-context adaptation module produces the in-context features, $\mathbf{x}_{ic}$, formally via:
\begin{equation}
    \mathbf{x}_{ic} = \textrm{LN}(\mathbf{x}) \cdot \mathbf{W}_{down} \cdot \mathbf{W}_{up}.
\end{equation}
Then, we can replace the original attention projector output with a feature that contains both pre-trained knowledge and task-specific knowledge through residual connections:
\begin{equation}
    \mathbf{\hat{Q}} = \textrm{AttentionProjector}_{q}(\textrm{LN}(\mathbf{E})) + \textrm{IC}_{q}(\mathbf{E}),
\end{equation}
where $\textrm{IC}_{q}(\cdot)$ represents the in-context adaptation module of $\mathbf{E}$ corresponding to the $\textrm{AttentionProjector}_{q}(\cdot)$ that generate query $\mathbf{Q}$.
In this paper, we only assign in-context adaptation modules to the attention projectors that generate $\mathbf{Q}$ and $\mathbf{K}$.
For experimental analysis, please see Experimental section.
\begin{table}[]
    \centering
    \begin{tabular}{lcccc}
    \toprule
        Methods& R@1 & R@2 & R@4  &Prams~(M) \\
        \bottomrule
        Non-fine-tuned	&72.5	&82.6	&89.1 &0.0\\
        \hline
         Fine-tuning&82.7  &88.7 &92.5  &85.6\\
         Attention Projector&82.7  &89.0 &92.8   &21.5\\
         Output Projector&82.6 &88.8 &92.6  &7.3\\
         MLP&82.0 &88.9 &92.9 &56.8\\
    \bottomrule 
    \end{tabular}
    \caption{Fine-Tuning Preferences Analysis of ViT on the CUB-200-2011 dataset.}
    \label{tab:analysis}
\end{table}

\subsection{Discriminative Perceptual Transfer}
With the cooperation of discriminative perception, the fine-grained understanding ability of the ViT model is enhanced.
However, the extra visual foundation model is time-consuming and memory-demanding for retrieval evaluation. 
In the classification field~\cite{hinton2015distilling}, network distillation has been shown to be one of the solutions to this problem. Inspired by network distillation, we propose a discriminative perceptual transfer to extend the knowledge distillation theory to retrieval tasks to achieve an efficiency-performance tradeoff.
Specifically, we separate discriminative perception into a separate branch, exploit a category proxy learning strategy to learn discriminative category proxy as carriers of discriminative knowledge, and transfer their discriminative knowledge.

We encode both the discrimination and distribution of labelled instances via proxy-guided learning~\cite{proxy}.
Then, for the retrieval embedding $\textbf{E}^{R}$ , we compute the proxy learning loss as follow:
\begin{equation}
    \mathcal{L}_{proxy} = -\textrm{log}\left( \frac{\textrm{exp}(-d(\Vert\textbf{E}^{R}\Vert, \Vert\textbf{c}\Vert))}{\sum_{\textbf{c} \in \textbf{C}}\textrm{exp}(-d(\Vert\textbf{E}^{R}\Vert, \Vert\textbf{c}\Vert))}\right),
\end{equation}
where $d(\Vert\textbf{E}^{R}\Vert, \Vert\textbf{c}\Vert)$ represents the distance between $\Vert\textbf{E}^{R}\Vert$ and $\Vert\textbf{c}\Vert$, $\textbf{c}$ denotes the category proxy corresponding to retrieval embedding $\textbf{E}^{R}$, $\textbf{C}$ is the category proxy set of the discriminative perception images $\mathbf{I}_d$, and $\Vert \cdot \Vert$ denotes the $L^2$-Norm.
\begin{table*}[t]
    \centering
    \begin{tabular}{ccccccccc}
    \toprule
         &\multicolumn{4}{c}{CUB-200-2011} &\multicolumn{4}{c}{Stanford Cars} \\
        \cmidrule(lr){2-5}\cmidrule(lr){6-9}
         \multirow{-2}{*}{Method} &R@1 &R@2 &R@4 &R@8 &R@1 &R@2 &R@4 &R@8 \\
         \midrule
         Proxy-Anchor~\cite{proxy}  &80.4 &85.7 &89.3 &92.3 &77.2 &83.0 &87.2 &90.2 \\ 
         HIST~\cite{hist} &75.6 &83.0 &88.3 &91.9 &\textbf{89.2} &\textbf{93.4} &95.9 &97.6 \\
        PNCA++ ~\cite{pnca++} &80.4 &85.7 &89.3 &92.3 &86.4 &92.3 &\underline{96.0} &\underline{97.8} \\
        \midrule
        Hyp-ViT~\cite{Hyp-ViT}  &84.2 &91.0 &94.3 &96.0 &76.7 &85.2 &90.8 &94.7 \\
        \textbf{DVA~(Ours)}  &\textbf{85.2} &\textbf{91.4} &\textbf{94.6} &\textbf{96.2} &\underline{88.0}&\underline{93.2} &\textbf{97.0} &\textbf{98.5} \\
         \bottomrule
    \end{tabular}
     \caption{Comparison with state-of-the-art methods in the closed-set setting on CUB-200-2011 and Stanford Cars. The best result is shown in \textbf{bold}, and the second-best result is \underline{underlined}.}
    \label{tab:table2_close_set}
\end{table*}

With the discriminative category proxy set $\textbf{C}$, and the retrieval embeddings $\hat{\textbf{E}}^{R}$ of the origin training images are used for distillation:
\begin{equation}
    \mathcal{L}_{dpt} = -\textrm{log}\left( \frac{\textrm{exp}(-d(\Vert\hat{\textbf{E}}^{R}\Vert, \Vert\textbf{c}\Vert))}{\sum_{\textbf{c} \in \textbf{C}}\textrm{exp}(-d(\Vert\hat{\textbf{E}}^{R}\Vert, \Vert\textbf{c}\Vert))}\right).
\end{equation}
After optimizing $\mathcal{L}_{dpt}$, the image encoder adapts to capture subtle inter-category differences transferred from the discriminative branch. This enhancement enables effective retrieval of visually similar items without relying on the computationally expensive discriminative perception module.

\subsection{Overall Function}
As illustrated in Fig.~\ref{fig:overview}, our objective function comprises two components: the proxy learning loss $\hat{\mathcal{L}}_{proxy}$ and the distillation loss $\mathcal{L}_{dpt}$. Finally, the total loss for our framework can be defined as:
\begin{equation}
 \begin{split}
     \hat{\mathcal{L}}_{proxy} &= -\textrm{log}\left( \frac{\textrm{exp}(-d(\Vert\textbf{E}^{R}\Vert, \Vert\textbf{c}\Vert))}{\sum_{\textbf{c} \in \hat{\textbf{C}}}\textrm{exp}(-d(\Vert\textbf{E}^{R}\Vert, \Vert\textbf{c}\Vert))}\right), \\ 
     \mathcal{L} &=  \hat{\mathcal{L}}_{proxy}+ \beta \mathcal{L}_{dpt},
    \label{loss}
    \end{split}
\end{equation}
where $\hat{\textbf{C}} \in\{\textbf{C}^{o},\textbf{C}^\mathrm{b}\}$ contains original training categories $\textbf{C}^{o}$ and background category $\textbf{C}^\mathrm{b}$, and $\beta$ is the hyper-parameter to weight the $\mathcal{L}_{dpt}$.
\section{Experiments}~\label{experiment}
\subsection{Experiment Setup}
\subsubsection{Datasets.} 
\textbf{CUB-200-2011}~\cite{cub} comprises 11, 788 bird images from 200 bird species. In the closed-set setting, the dataset is divided into training and testing subsets comprising 5,994 and 5,794 images, respectively, out of a total of 11,788 images. For the open-set setting, we employ the first 100 subcategories (comprising 5,864 images) for training, and the remaining subcategories (comprising 5,924 images) are used for testing.

\textbf{Stanford Cars}~\cite{stanfordcar} consists of 16,185 images depicting 196
car variants. Similarly, these images were split into 8,144 training images and  8,041 test images in the closed-set setting. For the open-set setting, we utilize the first 98 subcategories (comprising 8,054 images) for training and the remaining 98 subcategories (comprising 8,131 images) for testing.

\textbf{Stanford Dogs}~\cite{dataset2011novel} contains 20,580 images showcasing dogs across 120 subcategories. We use the 120 subcategories with 12,000 images for training and the remaining 120 subcategories with 8580 images for testing for the closed-set setting. For the open-set setting, we utilize the first 60 subcategories (comprising 10,651 images) for training and the remaining 60 subcategories (comprising 9,929 images) for testing.

\subsubsection{Evaluation protocols.} To evaluate retrieval performance, we adopt Recall@K with cosine distance in previous work~\cite{recallk}, which calculates the recall scores of all query images in the test set.
For each query image, the top \textbf{K} similar images are returned. A recall score of 1 is assigned if at least one positive image among the top \textbf{K} images; otherwise, it is 0.

\subsubsection{Implementation details.}
In experiments, we employ the ViT-B-16~\cite{vit} pre-trained on ImageNet21K~\cite{imagenet} as our image encoder.
All input images are resized to 256 $\times$ 256, and crop them into 224 $\times$ 224. In the training stage, we utilize the Adam optimizer with weight decay of 1e-4, and employ cosine annealing as the optimization scheduler.
The learning rate for all datasets is initialized to 1e-1 except for Stanford Dogs, which has a learning rate of 1e-2. The number of training epochs for all datasets is set to 10 and the batch size is set to 32.
We train our model on a single NVIDIA 3090 GPU to accelerate the training process.
\subsection{Comparison with State-of-the-Art Methods} \label{compare}
\begin{table*}
    \centering
    \begin{tabular}{ccccccccc}
    \toprule
         &\multicolumn{4}{c}{CUB-200-2011} & \multicolumn{4}{c}{Stanford Cars} \\
        \cmidrule(lr){2-5}\cmidrule(lr){6-9}
         \multirow{-2}{*}{Method} &R@1 &R@2 &R@4 &R@8 &R@1 &R@2 &R@4 &R@8\\
         \midrule
         DAS~\cite{das} &69.2 &79.3 &87.1 &92.6 &87.8 &93.2 &96.0 &97.9 \\
         IBC~\cite{ibc} &70.3 &80.3 &87.6 &92.7 &88.1 &93.3 &96.2 &98.2 \\
         Proxy-Anchor~\cite{proxy}  &71.1 &80.4 &87.4 &92.5 &88.3 &93.1 &95.7 &97.0 \\
         HIST~\cite{hist}  &71.4 &81.1 &88.1 &- &89.6 &93.9 &96.4 &- \\
        PNCA++~\cite{pnca++} &70.1 &80.8 &88.7 &93.6 &90.1 &94.5 &97.0 &98.4 \\
        FRPT~\cite{frpt} &74.3 &83.7 &89.8 &94.3 &\textbf{91.1} &\textbf{95.1} &\textbf{97.3} &\textbf{98.6} \\
        \midrule
        DFML~\cite{dfml} &79.1 &86.8 &- &- &89.5 &93.9 &- &- \\
        Hyp-ViT~\cite{Hyp-ViT} &\underline{84.0} &\underline{90.2} &\underline{94.2} &\underline{96.4} &86.0 &91.9 &95.2 &97.2 \\
         \textbf{DVA~(Ours)} &\textbf{84.9} &\textbf{90.6} &\textbf{94.5} &\textbf{96.7} &\underline{90.7} &\underline{94.8} &\underline{97.1} &\underline{98.4} \\
         \bottomrule
    \end{tabular}
     \caption{Comparison with state-of-the-art methods in the open-set setting on CUB-200-2011 and Stanford Cars.}
    \label{tab:table2}
    
\end{table*}

\begin{table*}
    \centering
    \begin{tabular}{ccccccccc}
    \toprule
         &\multicolumn{4}{c}{Open-set Setting} &\multicolumn{4}{c}{Closed-set Setting} \\
        \cmidrule(lr){2-5} \cmidrule(lr){6-9}
         \multirow{-2}{*}{Method} &R@1 &R@2 &R@4 &R@8 &R@1 &R@2 &R@4 &R@8\\
         \midrule
         Proxy-Anchor~\cite{proxy} &86.1 &91.9 &95.7 &97.6 &83.7  &89.9  &93.9  &96.5  \\
         HIST~\cite{hist} &86.2 &92.3 &95.6 &97.5 &84.7  &90.1  &93.4  &95.6  \\
        PNCA++ ~\cite{pnca++} &86.0 &92.3 &95.6 &97.7  &83.8  &90.1  &94.7  &96.8\\
        \midrule
        Hyp-ViT~\cite{Hyp-ViT}   &87.8 &92.8 &95.9 &97.6 &79.2  &86.9  &91.8  &95.2  \\
         \textbf{DVA~(Ours)} &\textbf{90.4} &\textbf{94.6} &\textbf{97.0} &\textbf{98.3} &\textbf{87.7}  &\textbf{93.0}  &\textbf{95.8}  &\textbf{97.3}  \\
         \bottomrule
    \end{tabular}
    \caption{Comparison with state-of-the-art methods on Stanford Dogs.}
    \label{tab:dog}
\end{table*}
\subsubsection{Closed-set Setting.}
We first compare our proposed DVA with previous competitive methods under closed-set setting. Quantitative comparison results for the CUB-200-2011 and Stanford Cars datasets are presented in Table~\ref{tab:table2_close_set}, and the results for the Stanford Dogs dataset can be found in Table~\ref{tab:dog}.
From these tables, it can be observed that our proposed {\name} outperforms other state-of-the-art methods on CUB-200-2011 and Stanford Dogs, and achieves competitive performance on Stanford Cars, which demonstrates the enhanced discriminative capability of our {\name} for fine-grained visual retrieval.
Specifically, in comparison with Hyp-ViT~\cite{Hyp-ViT}, the current state-of-the-art on CUB-200-2011, our DVA demonstrates a 1.0\% improvement in Recall@1.
The experimental results on the Stanford Cars indicate that our method outperforms the most of existing methods but falls slightly behind HIST~\cite{hist}.
We argue the possible reason may be attributed to the relatively regular and simpler shape of the cars.

\subsubsection{Open-set Setting.}
The open-set setting poses greater challenges compared to the closed-set setting due to the unknown subcategories in the test data.
The experimental results for the CUB-200-2011 and Stanford Cars datasets are shown in Table~\ref{tab:table2}, while the results for the Stanford Dogs dataset are provided in Table~\ref{tab:dog}.
The results reveal a consistent trend in both open and closed-set settings: our method outperforms other state-of-the-art approaches on CUB-200-2011 and Stanford Dogs, while delivering competitive results on the Stanford Cars dataset.
To be specific, in comparison with Hyp-ViT~\cite{Hyp-ViT} on CUB-200-2011, our DVA exhibits a 0.9\% improvement in Recall@1.
Experimental results on the Stanford Cars dataset show that our method outperforms most existing methods but lags behind FRPT~\cite{frpt} by a slight margin.
We attribute FRPT’s performance gains to its carefully designed but computationally expensive modules. In contrast, our method captures subtle differences through dual-vision adaptation while maintaining computational efficiency.
The results of experiments on Stanford Dogs are shown in Table~\ref{tab:dog}.
Our {\name} demonstrates a clear performance advantage over Hyp-ViT, further indicating its superior generalizability.

\subsection{Ablation Studies and Analysis}
\begin{table}[]
    \centering
    
    \begin{tabular}{lcccc}
    \toprule
     Setting &R@1 &R@2 &Latency~(ms)\\
          \midrule
          Base  &72.5 &82.6 &1.51\\
          Base + \textsc{ICA}  &83.6 &90.3  &1.51\\
          Base + \textsc{ICA}  + \textsc{OPA} &86.8 &92.0   &3.65\\
          Base + \textsc{ICA}  + \textsc{DP} + \textsc{DPT} &84.1 &90.2   &1.51\\
          Base + \textsc{ICA}  + \textsc{OPA} + \textsc{DPT} &84.9 &90.6 &1.51\\
         \bottomrule
    \end{tabular}
    \caption{The Recall@K results~(\%) of component ablation study on CUB-200-2011.}
    \label{tab:ablation}
\end{table}

\noindent
\textbf{Efficacy of various components.}
The proposed DVA comprises three essential components: Object-Perceptual Adaption~(\textsc{OPA}), In-Context Adaptation~(\textsc{ICA}), Discriminative Perception~(\textsc{DP}) of the \textsc{OPA}, and Discriminative Perceptual Transfer~(\textsc{DPT}).
We conducted ablation experiments on these components, and the results are reported in Table~\ref{tab:ablation}, with the Base representing the pure ViT.
The introduction of \textsc{ICA} improves the Recall@1 accuracy by 11.1\% on CUB-200-2011, which shows that ICA effectively guides the FGIR task to the task solved in the original process through feature adaptation.
\textsc{OPA} further improves the ability of the pre-trained model to distinguish similar subcategories on FGIR, bringing a gain of 3.2\% Recall@1 on CUB-200-2011, but it brings huge computational overhead.
\textsc{DPT} effectively removes the huge computational overhead brought by OPA and achieves a balance between performance and retrieval efficiency.
In particular, when only \textsc{DP} is used but no Object Perception, the performance decreases.
This shows that Object Perception effectively helps DVA perceive objects and thus better distinguish similar objects.

\begin{table*}[]
    \centering
    
    \begin{tabular}{lccccc}
    \toprule
    &\multicolumn{2}{c}{CUB-200-2011} &\multicolumn{2}{c}{Stanford Cars} \\
        \cmidrule(lr){2-3}\cmidrule(lr){4-5}
         \multirow{-2}{*}{Setting} &Recall@1 &Recall@2 &Recall@1 &Recall@2 &\multirow{-2}{*}{Prams~(M)} \\
         \midrule
          $\textbf{IC}_{q}$  &82.7 &89.1 &87.2	&92.9 &0.295\\
          $\textbf{IC}_{k}$ &81.8 &88.8 &85.6 &91.7 &0.295\\
          $\textbf{IC}_{v}$ &84.0 &90.0 &86.4	&92.3 &0.295\\
          $\textbf{IC}_{q}$, $\textbf{IC}_{k}$  &84.9 &90.6 &90.7	&94.8 &0.590\\
          $\textbf{IC}_{q}$, $\textbf{IC}_{v}$  &84.5 &90.5 &90.3	&94.5 &0.590\\
          $\textbf{IC}_{k}$, $\textbf{IC}_{v}$  &82.6 &89.0 &89.5	&94.1 &0.590\\
          $\textbf{IC}_{q}$, $\textbf{IC}_{k}$, $\textbf{IC}_{v}$  &84.4 &90.2 &90.4	&94.6 &0.885\\
         \bottomrule
    \end{tabular}
    \caption{The Recall@K results~(\%) of component ablation study on CUB-200-2011 and Stanford Cars.}
    \label{tab:ict}
\end{table*}

\noindent
\textbf{Analysis of In-Context Adaptation.} \label{ica}
To validate the effectiveness of the proposed ICA components, we further divide the \textsc{ICA} module into three parts: $\textbf{IC}_{q}$, $\textbf{IC}_{k}$, and $\textbf{IC}_{v}$, corresponding to the Q, K, and V projectors in the Attention Projector, respectively.
We perform ablation studies on these components, and the results are presented in Table~\ref{tab:ict}.
The results show that using $\textbf{IC}_{q}$ alone or in combination with $\textbf{IC}_{k}$ and $\textbf{IC}_{v}$ leads to better performance. Notably, the best results are achieved when $\textbf{IC}_{q}$ and $\textbf{IC}_{k}$ are used together. Therefore, we adopt this combination to construct the final ICA module in our {\name}.

\begin{figure}[htbp]
    \centering
\includegraphics[width=0.9\linewidth]{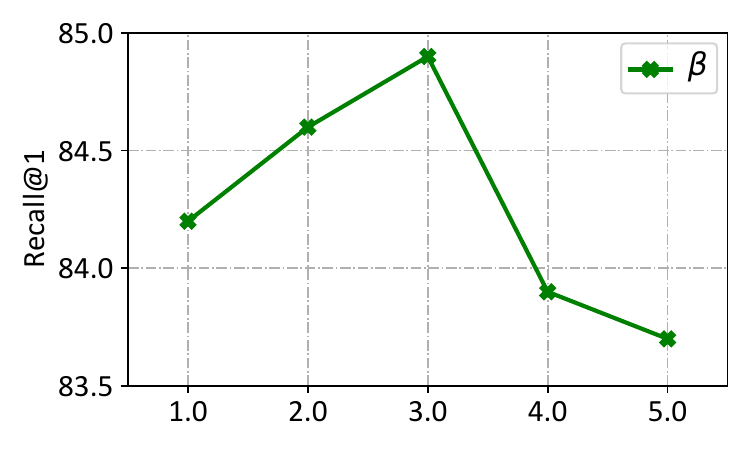}
    \caption{Analyses of hyper-parameter $\beta$ on CUB-200-2011.}
    \label{fig:beta}
\end{figure}

\noindent
\textbf{Sensitivity analysis of parameter $\beta$.}
As illustrated in Fig.~\ref{fig:beta}, we investigate the impact of varying values of $\beta$ in Eq.~(\ref{loss}).
The performance gradually increases with the value but decreases when the value exceeds 3.
The drop in performance can be attributed to the fact that as the value of $\beta$ increases, the learned feature space tends to receive samples adapted by the OPA module, while the input at inference is not processed by the OPA, resulting in a difference in the input space that causes performance degradation.
Consequently, we set $\beta$ to 3 for all datasets.

\subsubsection{Visualization.}
\begin{figure}
    \centering
\includegraphics[width=0.75\linewidth]{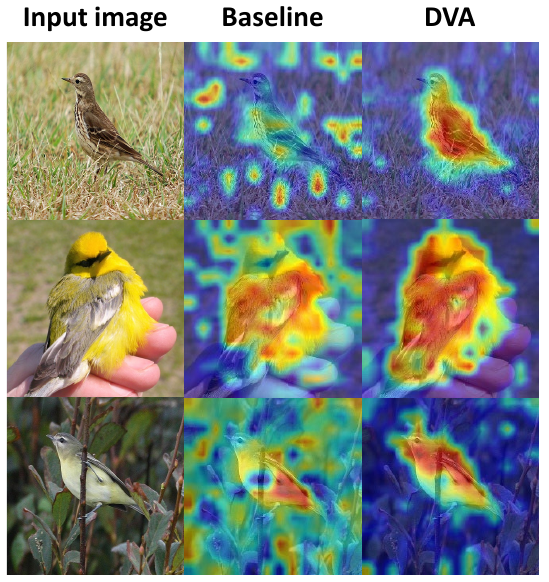}
    \caption{Class activation visualizations on CUB-200-2011. For each sample, we show the input image, the class activation map of the baseline model, and the proposed DVA.}
    \label{fig:visual}
\end{figure}

To explore the superiority of the designed DVA, We also conduct visualization as shown in Fig.~\ref{fig:visual}. 
The first column displays the input images. The second column shows the class activation maps generated by GradCAM for the baseline model, which processes the input images.
The third column presents the class activation maps for DVA using the same input. The results indicate that DVA effectively reduces attention to background regions and improves the baseline model’s focus on critical areas, capturing finer details in subcategories such as the head, wings, and tail.

% \begin{table}[]
%     \centering
%     \begin{tabular}{lcc}
%     \toprule
%      Methods &R@1\\
%           \midrule
%           \rowcolor{mygray}Fine-tuning ViT  &82.7 \\
%           \hline
%           CLIP-Adapter~\cite{gao2024clip}  &69.8  \\
%           CLIP-LoRA~\cite{cliplora}	&76.5 \\   AdaptFormer~\cite{chen2022adaptformer}  &73.3  \\
%           \textbf{DVA~(Ours)} &84.9 \\
%          \bottomrule
%     \end{tabular}
%     \caption{The Recall@K results~(\%) of different parameter efficient fine-tuning methods on CUB-200-2011.}
%     \label{tab:peft}
% \end{table}

% \subsubsection{Comparison with other efficient parameter fine-tuning methods.}
% We compare our method with two efficient parameter fine-tuning approaches: CLIP-Adapter~\cite{gao2024clip}, CLIP-LoRA~\cite{cliplora} and AdaptFormer~\cite{chen2022adaptformer}.
% %
% As shown in Table~\ref{tab:peft}, both methods perform well on the CUB-200-2011 dataset. However, our DVA still achieves superior results, demonstrating its ability to capture subtle differences and accurately distinguish similar subcategories through dual-vision adaptation.
% %
% Furthermore, DVA outperforms the fully fine-tuned ViT. This is expected, as fine-tuning a pre-trained model on a small fine-grained dataset often causes overfitting, reducing its generalization capability.
\section{Conclusion}
In this paper, we present Dual Visual Adaptation (DVA), a novel approach designed to alleviate overfitting in fine-grained image retrieval by retaining knowledge from pre-training.
DVA integrates three key components: Object-Perceptual Adaptation, which introduces a new background category and enhances discriminative object regions using a visual foundation model; In-Context Adaptation, which employs lightweight parameters to adapt fine-grained features within a frozen backbone; and Discriminative Perceptual Transfer, which distills discriminative knowledge into the encoder for efficient inference.
Extensive experiments on three fine-grained benchmarks demonstrate that DVA achieves strong performance with simple yet effective dual-vision adaptation.
% with minimal trainable parameters.
% In this paper, we propose a method called Dual Visual Adaptation (DVA), which aims to mitigate the risk of overfitting in traditional fine-grained image retrieval caused by full-model fine-tuning, by retaining knowledge from pre-training.
% %
% DVA introduces Object-Perceptual Adaption (OPA) and In-Context Adaptation (ICA) to guide a frozen pre-trained vision model in performing fine-grained retrieval tasks.
% %
% Specifically, OPA leverages a visual foundation model to create a new background category and enhance pixels contributing to category prediction, and helping the model better perceive critical objects and their distinguishing features. 
% %
% ICA introduce a small set of in-context adaptation parameters that enables the frozen pre-trained model to capture discriminative details through feature adaptation.
% %
% Additionally, DVA achieves a balance between discriminative perception and retrieval efficiency with the design of Discriminative Perceptual Transfer~(DPT). 
% %
% Extensive experiments on three fine-grained benchmarks show that our DVA achieves excellent performance with very few learnable parameters.
% Extensive experiments show that DVA has fewer learnable parameters and performs well on three in-distribution and three out-of-distribution fine-grained datasets, especially outperforming the state-of-the-art methods on the out-of-distribution datasets.
% \input{AnonymousSubmission/LaTeX/sect/appendix}

\bibliography{aaai2026}

\end{document}